%% file: root.tex
\documentclass[letterpaper, 10 pt, conference]{ieeeconf}  

\IEEEoverridecommandlockouts                              

\usepackage{graphicx} 
\usepackage{times} 
\usepackage{amsmath} 
\usepackage{amssymb}  
\usepackage[colorlinks=true, allcolors=black]{hyperref}
\usepackage[noend,linesnumbered,ruled]{algorithm2e}

\usepackage{xcolor}

\colorlet{todo}{red!70!black}

\title{\LARGE \bf
Productive Multitasking for Industrial Robots*
}

\author{
D. Wuthier$^1$,
F. Rovida$^2$,
M. Fumagalli$^1$,
and V. Kr{\"u}ger$^3$
\thanks{* This research was supported by the European Commission's Horizon 2020 Programme under grant agreement no.~723658 (SCALABLE) and by the Wallenberg Autonomous Systems Program (WASP) of Sweden.}
\thanks{$^1$ Automation and Control,
Department of Electrical Engineering,
Technical University of Denmark,
Elektrovej, Building 326,
2800 Kgs. Lyngby,
Denmark}
\thanks{$^2$ RiACT ApS,
A. C. Meyers V{\ae}nge 15,
2450 Copenhagen SV,
Denmark}
\thanks{$^3$ Robotics and Semantic Systems (RSS),
Department of Computer Science,
Lund University,
Ole R{\"o}mers V{\"a}g 3,
223 63 Lund,
Sweden,
\texttt{
volker.krueger@cs.lth.se}}
}

\begin{document}

\maketitle
\thispagestyle{empty}
\pagestyle{empty}

\begin{abstract}

  \input{tex/abstract.tex}

\end{abstract}

\begin{keywords}
industrial robots,
intelligent and flexible manufacturing,
robot skill,
behavior trees,
finite-state machines
\end{keywords}

\input{tex/introduction}

\input{tex/related_works.tex}

\input{tex/preliminaries}

\input{tex/preemption.tex}

\input{tex/implementation}

\input{tex/evaluation}

\input{tex/discussion.tex}

\bibliographystyle{IEEEtran}
\bibliography{references}

\end{document}

%% file: tex/abstract.tex
The application of robotic solutions to small-batch production is challenging:
economical constraints tend to dramatically limit the time for setting up new batches.
Organizing robot tasks into modular software components, called skills,
and allowing the assignment of multiple concurrent tasks to a single robot is potentially game-changing.
However, due to cycle time constraints,
it may be necessary for a skill to take over without waiting on another to terminate,
and the available literature lacks a systematic approach in this case.
In the present article, we fill the gap by
(a) establishing the specifications of skills that can be sequenced with partial executions,
(b) proposing an implementation based on the combination of finite-state machines and behavior trees, and
(c) demonstrating the benefits of such skills through extensive trials
in the environment of ARIAC (Agile Robotics for Industrial Automation Competition).


%% file: tex/introduction.tex
\section{Introduction}
\label{sec:introduction}

Highly dynamic markets challenge modern industries to absorb frequent changes in production
while keeping up with reasonable operation costs.
Robotics can potentially play a key role in addressing this issue,
but concerns of profitability severely limit the time ratio between recommission and exploitation.
This often results in highly repetitive tasks assigned to an expensive manual workforce.
However, letting industrial robots operate beyond safety fences
in order to collaborate with workers appears as a promising middle ground \cite{scalable}:
workers who are well aware of the situation and able to handle exceptions effectively
can combine their strengths with robots that are fast, accurate and restless.
Nevertheless, tapping into the full potential of this collaboration is still a matter of research.

An interaction between a worker and a robot can have two goals:
(a) programming tasks or (b) carrying out tasks collaboratively.
In the first case, a possible approach is \textit{robot skills}:
modular software components implemented once by robot experts
and deployed multiple times in various scenarios by the workers themselves \cite{pedersen2016robot}.
These skills consists of parametric blocks of executable code associated with pre- and post-conditions,
and they often refer to the verbs frequently occurring in standard operation procedures,
e.g., "pick", "place" and "insert".
This executable code is based on \textit{skill primitives} \cite{pedersen2013integration,krueger16PIEEE,krueger19rcim},
abstracting elementary sensing/perception capabilities and robot/tool commands,
e.g., "run camera", "localize object" and "move arm".
However, when it comes to performing tasks collaboratively,
the applicability of skills is limited by the necessity to complete their execution.
Consider the example of a worker $w$ assembling a product hand-in-hand with a robot $r$.
If $w$ notifies the presence of a faulty part,
$r$ should immediately reverse its execution \cite{laursen2018modelling};
if $r$ is becoming inaccurate at an operation,
it should put this operation on hold and get new instructions from $w$ \cite{steinmetz2016skill};
and if $r$ is in the process of clearing out the product and $w$ notices that he forgot a piece of fixture,
$r$ should bring the product back before releasing it on a conveyor system.
In this example, if $r$ was bound to terminate each skill it started,
then a faulty part could have damaged further parts,
an inaccurate movement could have resulted in a broken part,
and an erroneous assembly could have been conveyed to the next workstation.
\begin{figure}[t]
    \centering
    \includegraphics[width = 0.95\columnwidth]{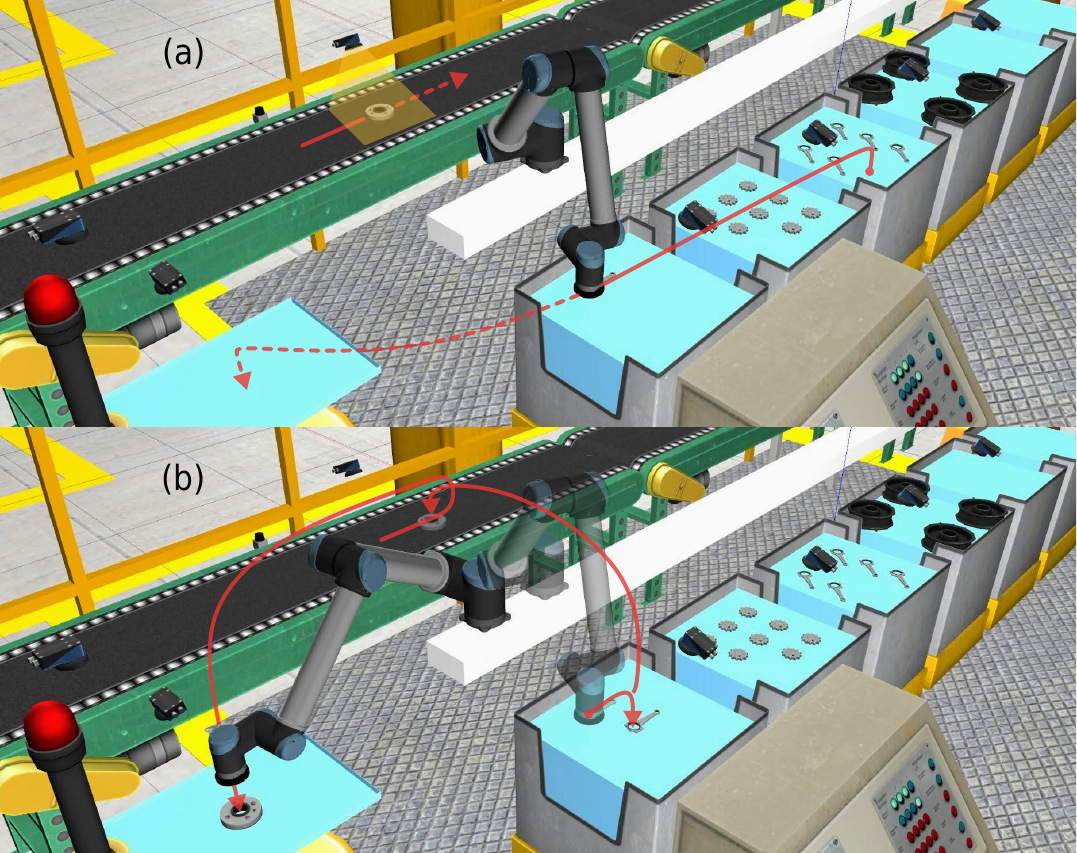}
    \caption{Environment of ARIAC (Agile Robotics for Industrial Automation Competition).
    (a) Execution of a \textit{place} skill in order to dispatch a piston rod on the tray of an AGV,
    while a disk is being detected on the conveyor belt.
    (b) Since this conveyor belt is running at a constant speed,
    waiting for the place skill to terminate would result in missing the opportunity to pick up the disk;
    the place skill is put on hold in order to start a \textit{pick} skill directed towards the disk.}
    \label{fig:preemption}
\end{figure}
In fact, the same limitation is highlighted by the recent Agile Robotics for Industrial Automation Competition (ARIAC)
\cite{aksu2018virtual} \footnote{\href{https://nist.gov/ariac}{https://nist.gov/ariac}},
where a robot (cf. Fig.~\ref{fig:preemption}) can be assigned with the tasks of
kitting parts on the delivery tray of an AGV from an array of bins ($T_A$) and
from a conveyor belt ($T_B$).
If a camera detects a disk on the conveyor belt (Fig.~\ref{fig:preemption}a)
and the robot is executing a \textit{place} skill $S$ in the context of $T_A$,
then completing the execution of $S$ before switching to $T_B$ could result in missing the opportunity of picking up the disk. 
By analogy to the design of operating systems,
$S$ is required to be \textit{preemptive} \cite{bach1986design}. 

A fundamental characteristic of preemptive skills is that
they maintain the system in a \emph{workable state},
a state from which the execution can be resumed by other skills.
In the present paper, we aim at establishing the following three contributions:
\begin{enumerate}
    \item we formally introduce a notion of preemption for the execution of robot skills;
    \item we introduce a joint BT-FSM model for use in robotics that combines Hierarchical Finite-State Machines (HFSMs) \cite{bohren2011towards}
    and Behavior Trees (BTs) \cite{colledanchise2018behavior}, we explore its use for modeling preemptive skills; and
    \item we demonstrate the benefits of preemptive skills versus non-preemptive ones
    through extensive trials in the ARIAC's simulation environment.
\end{enumerate}

The rest of this paper is structured as follows:
Sec.~\ref{sec:related_works} gives the context of our work,
Sec.~\ref{sec:preliminaries} introduces some important terms and concepts related to skill-based programming,
Sec.~\ref{sec:preemption} further formalizes the problem and specifies preemption for skills,
Sec.~\ref{sec:implementation} details our suggested implementation,
Sec.~\ref{sec:evaluation} describes an evaluation in the environment of ARIAC,
Sec.~\ref{sec:discussion} discusses the alternatives to our approach,
and concludes this paper and gives an outlook on future works.

%% file: tex/related_works.tex
\section{Related works}
\label{sec:related_works}

Primitives can assemble into a program using, e.g., Finite-State Machines (FSMs)
\cite{stolt2011force, wahrburg2015combined}.
FSMs are intuitive, but they do not handle concurrency explicitly
and they scale up unfavorably in more complex tasks
due to an exceedingly large amount of transitions. Hierarchical FSMs (HFSMs)
mitigate these problems using container states \cite{bohren2011towards, brunner2016rafcon},
and they fit well in the skill-based programming approach,
as one hierarchical level can be dedicated to tasks and another to skills \cite{thomas2013new}.
Steinmetz and Weitschat introduced a concept of preemption for skills based on HFSMs
\cite{steinmetz2016skill},
but it is mainly limited to integrated teaching strategies.
Preemption is a natural feature of Behavior Trees (BTs) \cite{colledanchise2018behavior}, with applications in skill-based programming \cite{rovida2017skiros}
and collaborative robotics \cite{paxton2017costar}.
Using BTs, skills are polled at a specific frequency with a possible preemption at every time step.
Recent video game development literature suggests that in terms of scalability and modularity, the combination of BTs and FSMs could be even more advantageous
\cite{anguelov2015separation, miyake2017character}.
With a concrete focus on preemption, Danga et al. discussed preemption of mobile robots on a task level, i.e., they can stop the task half way, but they did not consider preemption with a skill \cite{dang16CIRP} \cite{nielsen14conf}.

%% file: tex/preliminaries.tex
\section{Preliminaries}
\label{sec:preliminaries}

\begin{figure}[t]
\centering
\includegraphics[width = 0.9 \columnwidth]{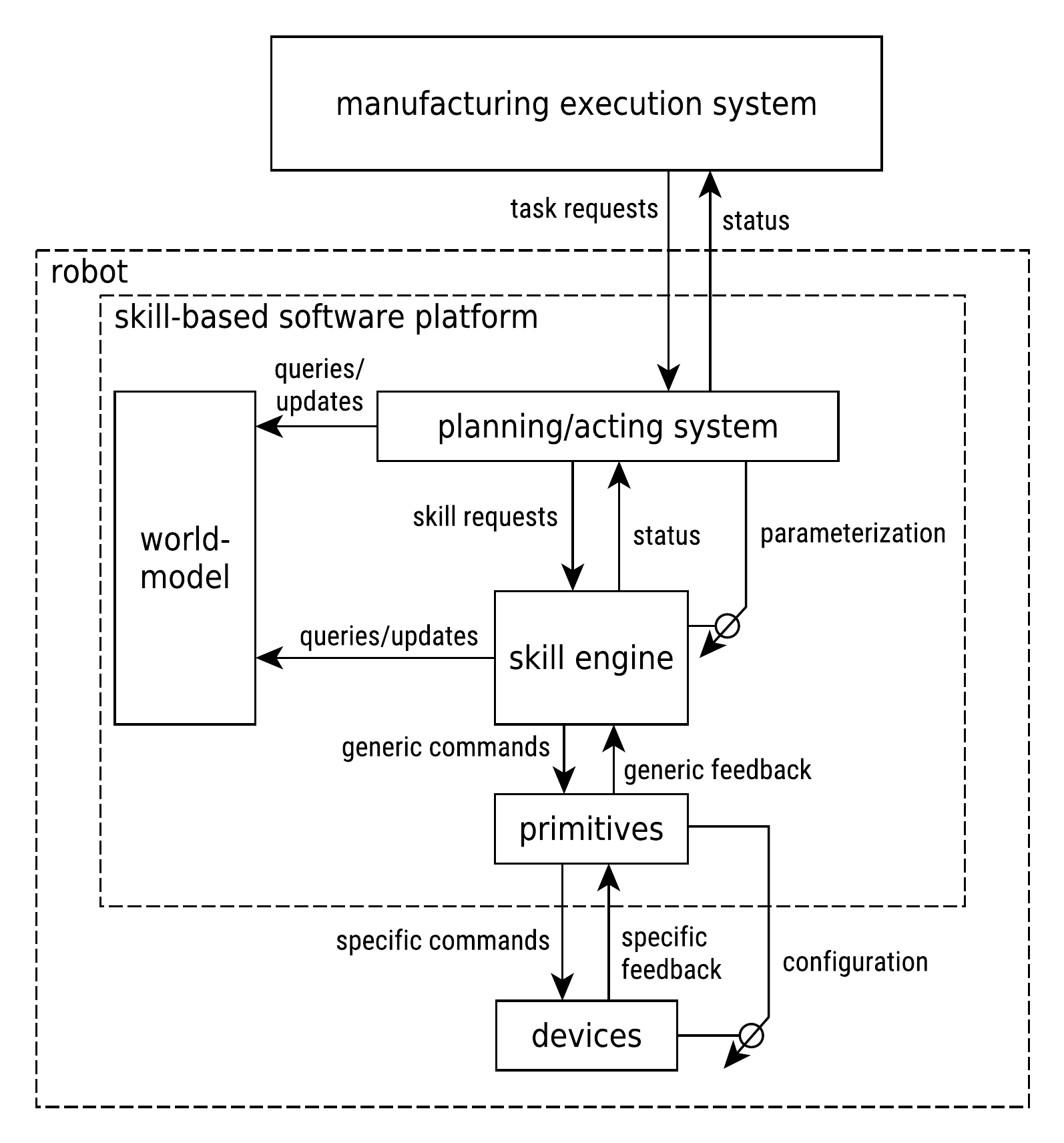}
\caption{Our architecture of skilled robot capable of processing the tasks assigned by a manufacturing execution system.
An arrow with a circle represents parameterization.}
\label{fig:architecture}
\end{figure}

\begin{figure*}
\centering
\includegraphics[width = 0.95 \textwidth]{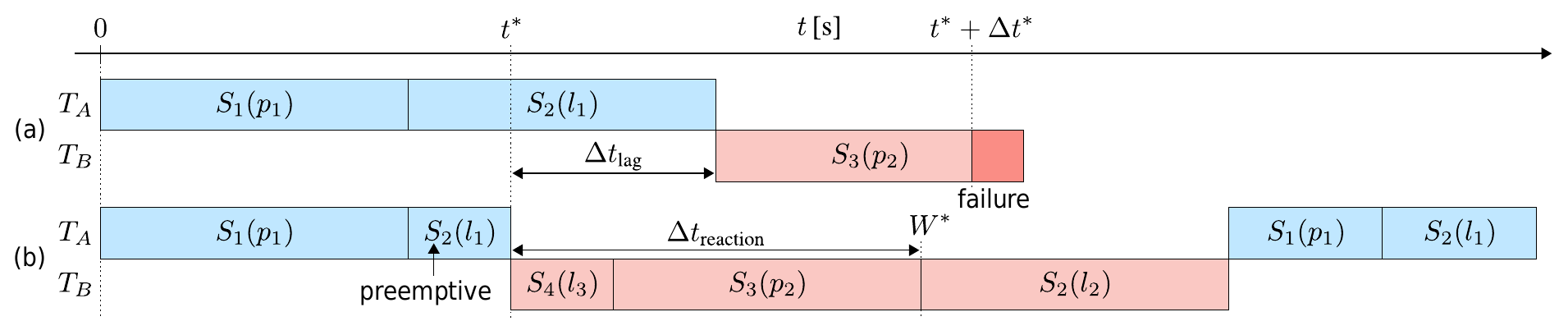}
\caption{Example of ARIAC in Sec.~\ref{sec:introduction} with parts $p_1$ (piston rod) and $p_2$ (disk);
placing locations $l_1$ (AGV's tray), $l_2$ (AGV's tray), and $l_3$ (intermediate location in bins);
tasks $T_A$ ($p_1$ is on $l_1$) and $T_B$ ($p_2$ is on $l_2$); and skills $S_1$ (pick in bins), $S_2$ (place on AGV), $S_3$ (pick on conveyor) and $S_4$ (place in bins).
A time constraint $\Delta t^*$ (the amount of time $p_2$ remains reachable) appears at $t^*$ (when $p_2$ becomes visible) on the attainment of a world-state $W^*$ (when $p_2$ is held in the robot's end-effector).
(a) Scenario with non-preemptive skills: after $t = t^*$, $S_2(l_1)$ runs until completion for a duration of $\Delta t_{\text{lag}}$,
causing $S_3(p_2)$ to fail.
(b) Scenario with preemptive skills: $S_4(l_3)$ preempts $S2(l_1)$ so that $S_3(p_2)$ executes with $\Delta t_{\text{reaction}} \leq \Delta t^*$, and $S_2(l_2)$ finalizes $T_B$. Then, $S_1(p_1)$ and $S_2(l_1)$ run once again to fulfill $T_A$.}
\label{fig:timing}
\end{figure*}

We clarify here some of the terms used throughout this paper
and then describe our architecture accordingly.
See \cite{ingrand2017deliberation} for general concepts related to deliberation in robotics
and \cite{ieeestd18722015} for a more exhaustive terminology of robotics and automation.

\subsection{Definitions}


\subsubsection{World-state}
Collection of elements, properties and relations that represents
a group of robots and their environment at some specific time $t$.

\subsubsection{Task}
Target world-state that is meaningful from a factory perspective,
e.g. a kitting task where a set of products of a certain type has to be collected on a delivery tray.

\subsubsection{Action}

Deterministic transition between two world-states,
including the physical effects of manipulation
and the non-physical effects of perception and information harvesting,
defined by sets of pre- and post-conditions, e.g.,
the transition from "object $x$ held in gripper $y$" to "object $x$ laying on location $z$".

\subsubsection{Primitive}

Software component that is specific to a certain model of device (or a certain implementation of an algorithm)
and that facilitates the reuse of executable robot codes
by exposing a uniform interface on all the devices (or algorithms) of the same type.
Primitives are the building blocks of some \textit{hardware-abstraction layer},
and they can be distinguished between:
\begin{itemize}
\item \textit{device primitives}, interfacing a sensor/actuator,
e.g. "acquire image from camera with specified exposition";
\item \textit{perception primitives}, interfacing a perception algorithm,
e.g. "locate an object on an image"; and
\item \textit{movement primitives}, interfacing a control algorithm,
e.g. "Move the arm linearly to a specified pose".
\end{itemize}

\subsubsection{Skill}

An executable robot code that exploits primitives to realize an action through a robot.
This code may be the result of a hierarchical composition
of other skills.


\subsection{Architecture}

We consider a production plant where a Manufacturing Execution System (MES) supervises the overall operations
by assigning tasks to the available robots.
In such a plant (cf. Fig.~\ref{fig:architecture}),
a robot is equipped with a \textit{skill-based software platform} \cite{krueger16PIEEE, krueger19rcim}
that processes the tasks requested by the MES using a planning/acting system.
This system parameterizes and coordinates the execution of skills by communicating with a \textit{skill engine}
that interfaces to the devices at hand through a set of primitives.
Both the planning/acting system and the skill engine 
query and update current believes about the world-state in a \textit{world-model}.

%% file: tex/preemption.tex
\section{Preemption}
\label{sec:preemption}


Consider the example of ARIAC in Sec.~\ref{sec:introduction} with parts $p_1$ (piston rod) and $p_2$ (disk);
placing locations $l_1$ (AGV's tray), $l_2$ (AGV's tray), and $l_3$ (intermediate location in bins);
tasks $T_A$ ($p_1$ is on $l_1$) and $T_B$ ($p_2$ is on $l_2$); and skills
$S_1$ (pick in bins), $S_2$ (place on AGV), $S_3$ (pick on conveyor) and $S_4$ (place in bins).
Fig.~\ref{fig:timing} 
represents the emergence of a time constraint $\Delta t^*$ (the amount of time $p_2$ remains reachable)  at time $t = t^*$ (when $p_2$ becomes visible) on the attainment of a world-state $W^*$
(when $p_2$ is held in the robot's end-effector):
while task $T_A$ is under execution, the robot is requested to complete task $T_B$ so that $W = W^*$ before $t = t^*+\Delta t^*$.
This constraint causes a change of priority between the tasks $T_A$ and $T_B$.
Generally speaking, let $[N]$ denote the set of integers $\{1, 2, \dotsc, N \}$,
let $\{S_i\}_{i = 1}^N$ with $N \in \mathbb{N}$ be the set of available skills,
and assume that the sequence
$\big(S_{a_i}(x_i)\big)_{i = 1}^M$ with $M \in \mathbb{N}$, $a_i \in [N]$, and sets of parameters $x_i$,
allows the fulfillment of $T_A$.
Then, if $T_B$ becomes more urgent than $T_A$ during the execution of $S_{a_m}(x_m)$ with $m \leq M$,
$S_{a_m}(x_m)$ would in principle take a duration of $\Delta t_{\text{lag}}$ to terminate
before letting another sequence
$\big(S_{b_i}(y_i)\big)_{i = 1}^J$ with $J \in \mathbb{N}$, $b_i \in [N]$, and sets of parameters $y_i$,
attempt to fulfill $T_B$ from that world-state at time $t^* + \Delta t_\text{lag}$.
Since $\Delta t^*$ constrains the attainment of a world-state $W^*$,
there is a risk that $\Delta t_{\text{lag}}$ causes $\Delta t^*$ to be violated.
In the example of Fig.~\ref{fig:timing}a, $T_A$ can be fulfilled by $\big(S_1(p_1), \, S_2(l_1)\big)$,
and $T_B$ by $(S_3(p_2), \, S_2(l_2))$, but $S_3(p_2)$ fails due to the completion of $S_2(l_1)$.


\subsection{Mechanism}

Tasks can be fulfilled by chaining the execution of skills \cite{pedersen2016robot,rovida2017skiros},
and for this to work, successful skill executions must be verified by checking post-conditions \cite{krueger16PIEEE}.
Furthermore, pre- and post-conditions are satisfied only in workable state \cite{krueger19rcim}.
Thus, a skill can only be executed if the world is in a workable state.
We return to a workable state with a large likelihood by letting skills terminate. 
An incomplete skill execution can leave the robot in a non-workable state where none of the skills can be executed.
Thus, the completion of tasks requires the entire execution of every skill.

Nevertheless, by devising a skill engine that abstracts away the alteration of a workable state
resulting from the partial execution of a skill,
other skills can still be started as if their pre-conditions were satisfied.
This way, tasks can be completed without requiring the entire execution of every skill,
and switches between tasks become instantaneous.
We refer to such a skill that can execute partially as \textit{preemptive},
and a request for the next skill to start before the previous one terminates as \textit{preemption}.
Generally speaking, when $S_{a_m}$ is preemptive,
the change of priority can be handled straightaway by starting a new sequence
$\big(S_{c_i}(z_i)\big)_{i = 1}^K$ with $K \in \mathbb{N}$, $c_i \in [N]$, and sets of parameters $z_i$,
that is derived \textit{on-the-fly} based on $T_B$ and the world state at time $t^*$.
Preemptive skills are therefore more likely to develop a reaction time $\Delta t_\text{reaction} \leq \Delta t^*$.
Back to Fig.~\ref{fig:timing}b, $S_4(l_3)$ preempts $S_2(l_1)$ in such a way that $T_A$ is "put on hold"
and $S_3(p_2)$ succeeds. Then, $T_B$ is finalized by $S_2(l_2)$
and $T_A$ is  "resumed" by executing $S_1(p_1)$ and $S_2(l_1)$ once again.

\subsection{Interpretations}

If $S_{c_1}(z_1)$ preempts $S_{a_m}(x_m)$ by carrying on with same sequence of primitives,
e.g. at the stage of withdrawing from an assembly,
then $c_{i + 1} = b_i$ $\forall i \in [J]$;
preemption can be regarded as a \textit{continuation} of the execution flow.
Otherwise, if the progression of a task can persist in some way after the execution of $S_{c_1}(z_1)$,
preemption could relate to the idea of \textit{putting on hold}.
Fig.~\ref{fig:preemption}b illustrates this example
with the headway of a part toward its destination remaining constant after a temporary placement in another bin.
Resuming and terminating this task therefore takes less time than completing it in one row.
In any other circumstances, the execution will be \textit{reversed} as investigated in
\cite{laursen2018modelling}.
The concept of preemption has therefore a spectrum of interpretation spanning
from "continue" to "reverse" through "put on hold".
However, it differs fundamentally from the notion of \textit{emergency stop},
which requires a manual intervention before any further execution of skills.

%% file: tex/implementation.tex
\section{Implementation}
\label{sec:implementation}

%

In this section, we propose a procedure for implementing preemptive skills in a systematic way.
It is based on our BT-FSM model that combines BTs and FSMs. We will introduce BT-FSM next.


\subsection{Joint BT-FSM model}
\label{sec:btfsm}

\begin{figure}[t]
\centering
\includegraphics[width = \columnwidth]{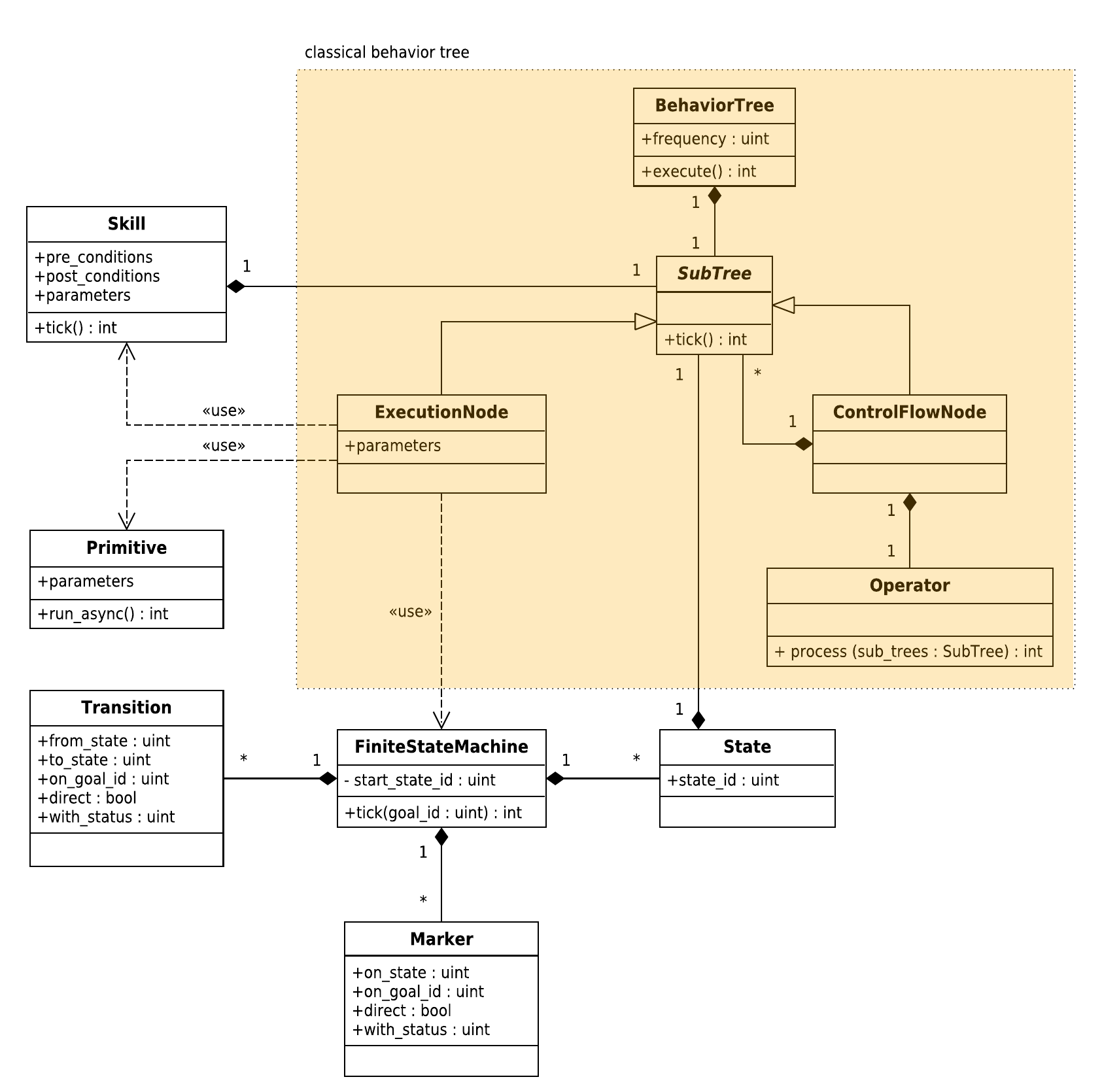}
\caption{UML class diagram of our Behavior Tree (BT) implementation integrating Finite-State Machines (FSMs).
This implementation extends the classical definition of BTs by allowing execution nodes
to drive a FSM toward multiple goals.
The use of primitives and skills by the execution nodes relates to the context of skill-based programming.}
\label{fig:uml}
\end{figure}

\begin{figure}[t]
\centering
\includegraphics[width = 0.95 \columnwidth]{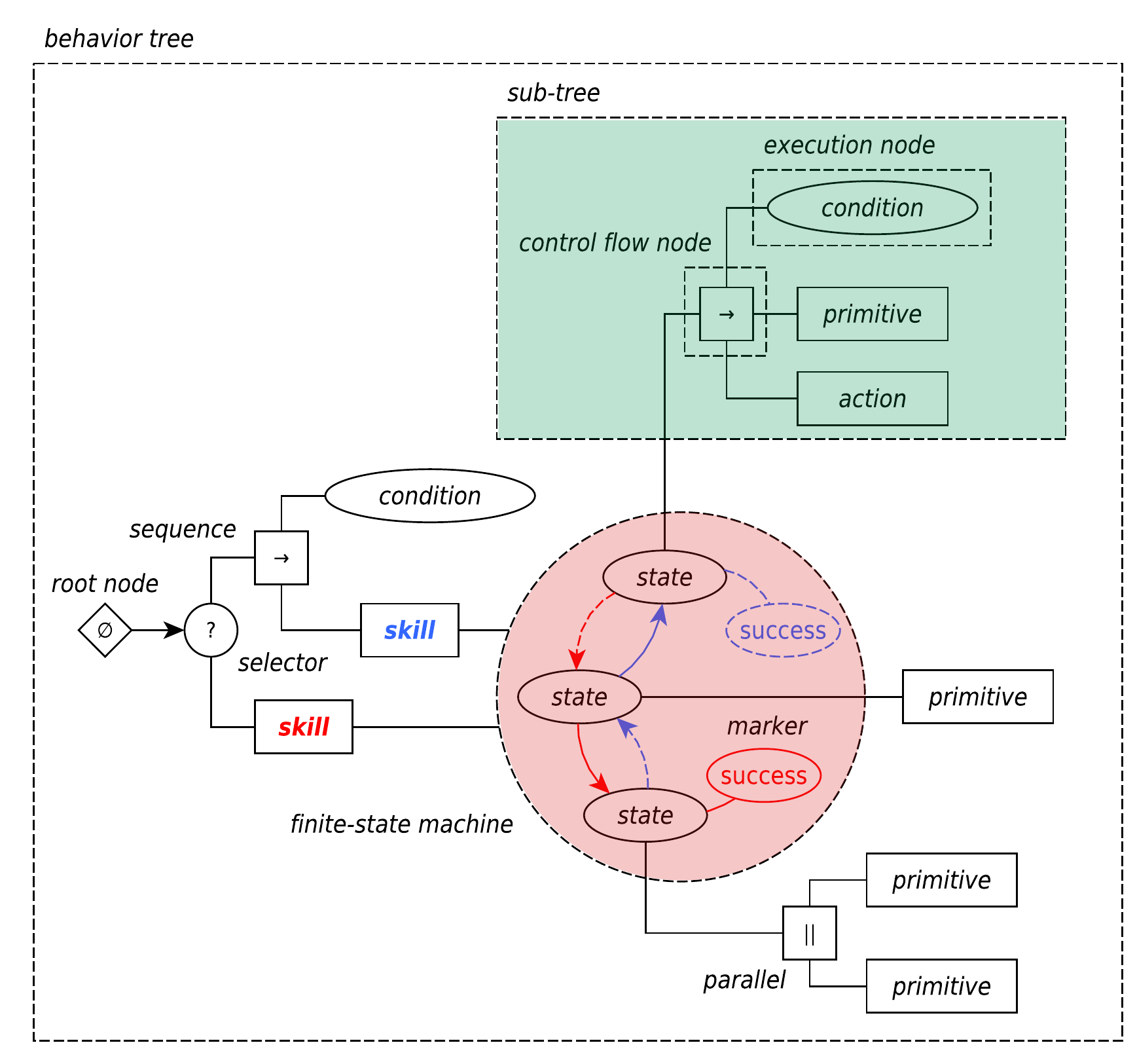}
\caption{Illustration of our implementation of behavior trees integrating finite-state machines
according to Fig.~\ref{fig:uml}.}
\label{fig:abstract}
\end{figure}


BTs are remarkable at providing the skill developer with a clear overview of the resulting behavior.
Also, they often preserve their clarity while scaling up
and they are useful at highlighting programming errors.
We implemented BTs as shown in the UML diagram of Fig.~\ref{fig:uml}.
The leaves consist of \textit{execution nodes} realized (dotted arrow) either by conditions or actions. In the context of skill-based programming,
actions are associated with skills and skill-primitives.
An execution node is an elementary instance (white arrow) of a \textit{sub-tree},
which hierarchically composes larger sub-trees using \textit{control-flow nodes}.
Control-flow nodes rely on (black dagger) operators such as \textit{sequence}, \textit{selector} or \textit{parallel}
(see \cite{colledanchise2018behavior})
in order to compute the outcome of such a hierarchical composition of sub-trees.
A sub-tree can also constitute (black dagger):
\begin{enumerate}
    \item a standalone BT, when associated with a specific polling (or \textit{ticking}) frequency;
    \item a skill, with pre- and post-conditions; and
    \item the state of a FSM, with an identifier.
\end{enumerate}
In general, \textit{a standalone BT represents the top-level behavior of an agent,
while a skill allows the reuse of a sub-tree on several branches}.



As depicted by Fig.~\ref{fig:abstract},
a FSM constituted by such states differs in this paper from traditional implementations:
it executes synchronously with the ticking of a BT,
and it can be driven toward multiple goals with possible changes of goal at any time
(here: the red and blue skills).
The edges representing the possible transitions from one state to the other
are associated with both conditions and goals,
and the same applies to the markers designating the possible terminations on a state.
The transitions and markers represented with full lines
are evaluated after polling the status of the current state,
whereas the ones represented with dashed lines are followed straightaway
(hence the flags \texttt{direct} on Fig.~\ref{fig:uml}).
The execution of such a FSM toward a specific goal is requested by an execution node.

\subsection{Procedure}

Equipped with our BT-FSM model, we now show how a BT-FSM can
implement preemptive skills with the following six systematic steps (yet to be taken manually):
\begin{enumerate}
    \item list the behaviors the robot should adopt,
    \item translate these behaviors into a BT that parameterizes and coordinates the execution of $N$ skills,
    \item determine which transitions between these skills can be requested by this BT
    (among $N^2$ possibilities; a transition from a skill to itself is related to a change of parameter),
    \item determine which of these transitions can occur before completion (i.e. with preemption),
    \item implement the execution of each skill independently as linear FSMs (sequences), and
    \item add all the transitions implied by the possible sequencing and preemptions of each skill.
\end{enumerate}
In principle, these steps are similar to those one has to take when designing a BT or a FSM.
As a concrete example, we apply this procedure for an ARIAC-Task in Sec.~\ref{sec:evaluation}.

%% file: tex/evaluation.tex

\section{Evaluation}
\label{sec:evaluation}
Based on the ARIAC example, we will first demonstrate how to design a BT-FSM for the preemption problem and then run it to demonstrate its use.

At the first glance,
the behavioral logic behind the scenario of Fig.~\ref{fig:preemption} seems simple.
However, as human, we are highly effective at dealing with such situations at an unconscious level,
and when it comes to programming a robot, every detail has to be considered explicitly.
That means (step 1):
\begin{itemize}
    \item monitoring the conveyor belt continuously in order to preserve an accurate belief
    about the queue of objects to be picked up;
    \item picking up objects on the conveyor if the queue is not empty but the gripper is;
    \item placing objects on a temporary location if the queue and the gripper are not empty,
    and if these objects are not sufficiently close from their final destination;
    \item picking up objects in the bins if the gripper and the queue are empty; and
    \item placing objects on the AGV if the queue is empty but not the gripper.
\end{itemize}

\subsection{Behavior tree}
\label{sec:shared}

\begin{figure}[t]
\centering
\includegraphics[width = 0.74\columnwidth]{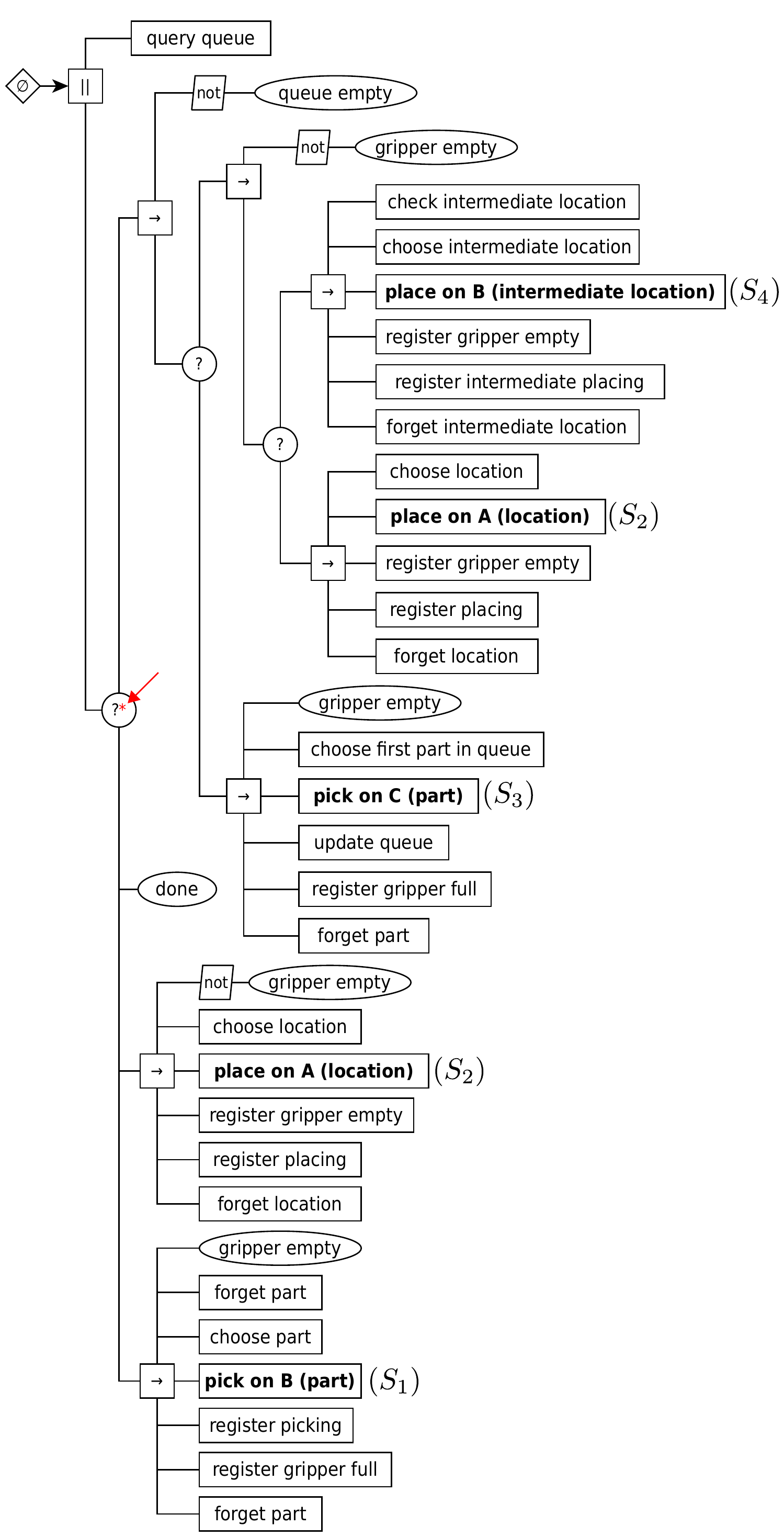}
\caption{Behavior tree modeling the robot's behavior in the ARIAC scenario.
Skills are designated by bold letters
and they apply to an area: the AGV (A), the bins (B) and the conveyor (C).
As explained in Sec.~\ref{sec:experiment},
the addition of a memory operator on the selector designated by the red arrow can inhibit preemptions.}
\label{fig:bt}
\end{figure}

\begin{figure}[t]
\centering
\includegraphics[width = 0.9\columnwidth]{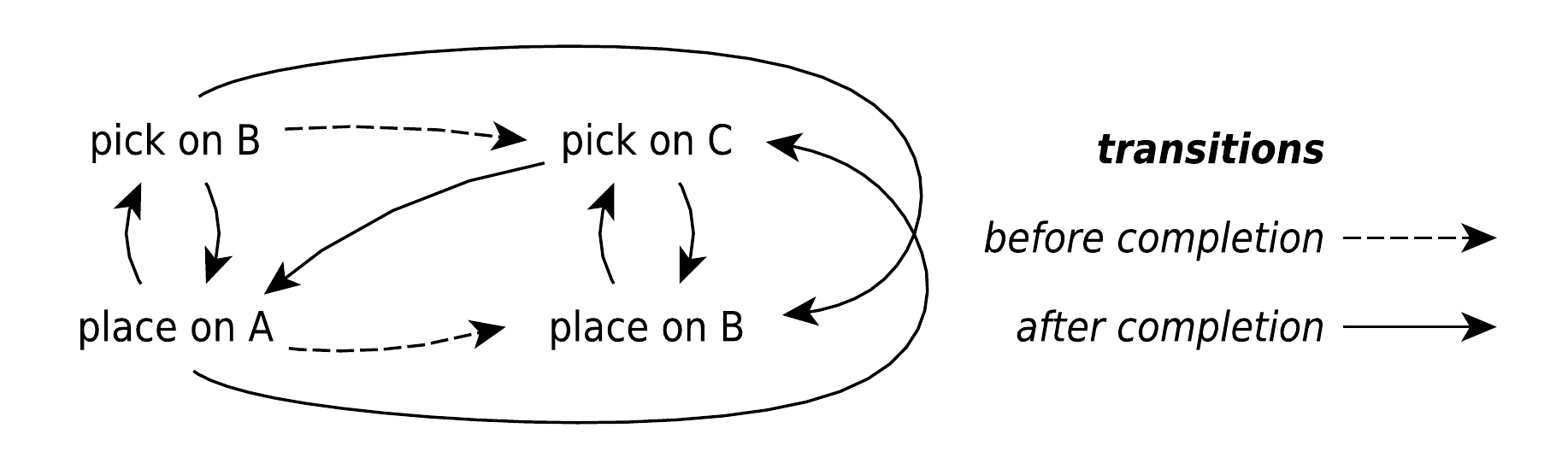}
\caption{Transitions requested by the behavior tree of Fig.~\ref{fig:bt}.
Dashed lines represent the transitions occurring before completion (i.e. \textit{preemption}),
and full lines, after completion (i.e. sequencing).}
\label{fig:skill_diagram}
\end{figure}

The BT of Fig.~\ref{fig:bt} implements this behavior in a conventional manner (step 2),
which can  qualify as being intuitive.
Its structure is based on a set of design patterns.
At the root, a parallel operator ensures the continuous monitoring of the conveyor.
Then, a decision tree pattern allows to distinguish the four cases relating to the conditions
"queue empty" and "gripper empty".
A condition is added in between in order to express the final goal
requiring the collection of a given amount of parts on the delivery tray of the AGV.
Each branch involves the execution of a skill (designated by bold letters),
and a sequence operator allows to run pre-condition checks beforehand and post-condition updates afterwards.
Let's denote the three areas around the robot, AGV, Bins, and Conveyor, as $A$, $B$ and $C$, respectively.
There are four skills:
\begin{enumerate}
    \item "pick from $B$" ($S_1$),
    \item "place on $A$" ($S_2$),
    \item "pick from $C$" ($S_3$) and
    \item "place on $B$ ($S_4$).
\end{enumerate}
Consequently, 16 different transitions can be requested.
However, an inspection of the BT of Fig.~\ref{fig:bt} (step 3) reveals that only 9 are necessary:
7 after completion and 2 before (step 4).
These transitions are represented on Fig.~\ref{fig:skill_diagram}.
Note how preemption appears as a natural feature of BTs.

The further use of BTs down the hierarchy in order to implement these skills can be cumbersome.
Assume that $S_1$ and $S_3$ are implemented with BTs all the way down to the primitive level.
Then, when $S_3$ is requested to preempt $S_1$,
the current state of the execution has to be stored in memory (either in the world-model or locally),
and $S_3$ has to evaluate numerous conditions in order to find a valid transition from the current state.
For example, if the robot is lifting a part, this part should be immediately put back, the vacuum gripper disabled,
and the arm withdrawn; if the part has not been grasped yet, the arm should simply withdraw, etc.
This pattern limits the readability of the BT because it generates many dependencies among the branches.
Since implementing preemptions is mostly about defining transitions, using a FSM here seems more appropriate to us: it would provide the transitions while keeping the readability of the BT.


\subsection{Finite-state machine}

\begin{figure}[t]
\centering
\includegraphics[width = 0.8\columnwidth]{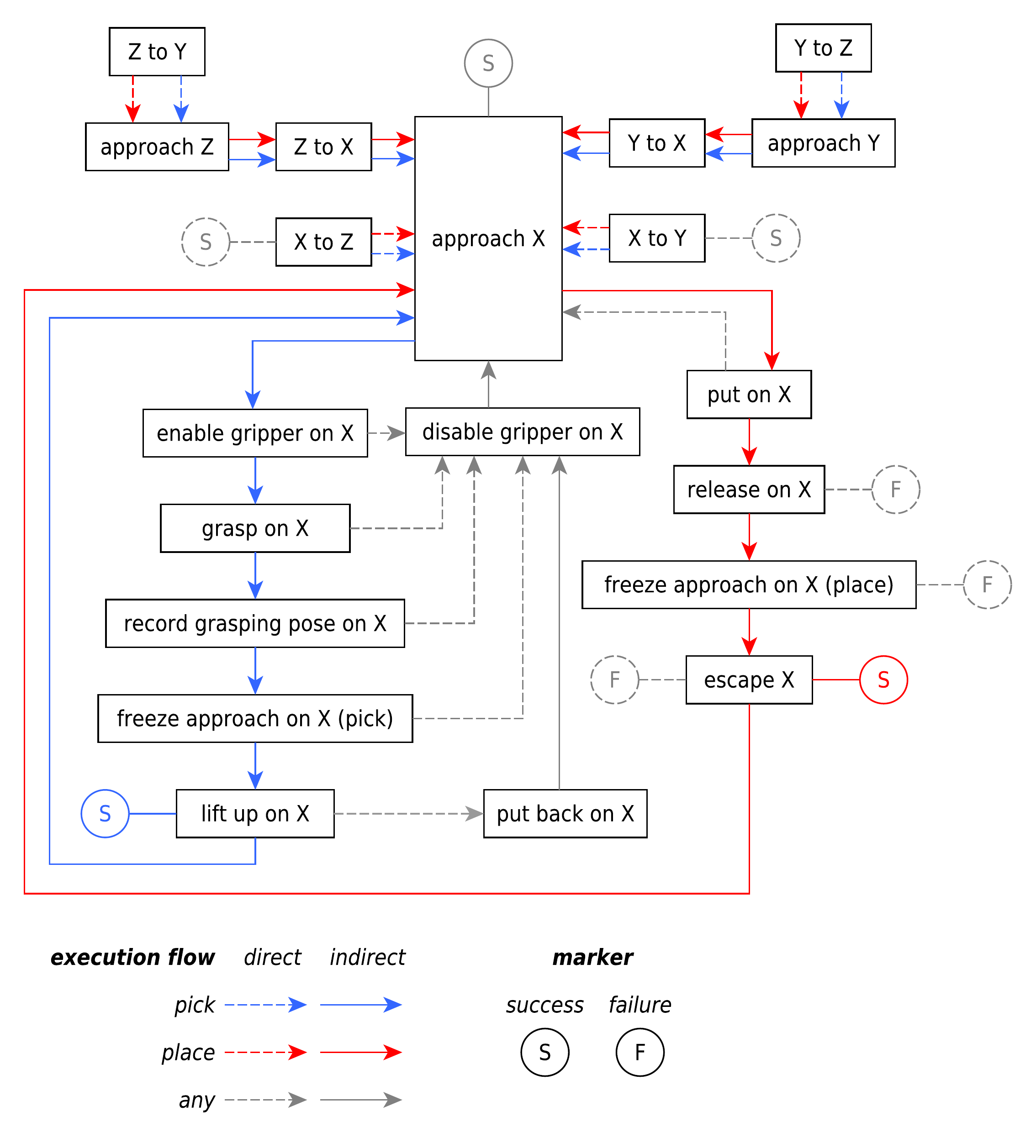}
\caption{Template of FSM for implementing the skills requested by the BT of Fig.~\ref{fig:bt} down to the primitive level. $X$, $Y$ and $Z$ denote areas.}
\label{fig:fsm}
\end{figure}

This FSM should implement the picking and the placing in each area, including all the necessary transitions.
We generated it based on the template shown on Fig.~\ref{fig:fsm}.
This template represents a function \texttt{generate\_fsm}:
$(X, Y, Z) \in \{A, B, C\}^3 \rightarrow (N, E, M)$
that generates a set of nodes $N$ (states), a set of edges $E$ (transitions) and a set of markers $M$
based on a permutation of the three areas.
We obtain the overall FSM with the union 
$
\left(
\bigcup_{i \in [3]} N_i,
\bigcup_{i \in [3]} E_i,
\bigcup_{i \in [3]} M_i
\right)
$
with
\[
\begin{aligned}
& \quad (N_1, E_1, M_1) = \text{generate\_fsm}(A, B, C), \\
& \quad (N_2, E_2, M_2) = \text{generate\_fsm}(B, C, A), \\
& \quad (N_3, E_3, M_3) = \text{generate\_fsm}(C, A, B). \\
\end{aligned}
\]

With regard to Fig.~\ref{fig:fsm}, the picking sequence is implemented with the blue transitions,
and the placing sequence with the red ones (step 5).
The grayed transitions correspond to a control flow that is followed
when no transition associated with the requested skill is found,
or when the part to pick (or location where to place) is changed (step 6).
The transit between the three areas is implemented on the top.

\subsection{Testing the BT-FSM for Preemption}
\label{sec:experiment}


The two possible preemptions of Fig.~\ref{fig:skill_diagram} are representative of our approach.
By leaving them unused, we can establish a baseline for preemptive skills.
To do this, we use a trick by simply
adding a memory operator on the selector designated by the red arrow on Fig.~\ref{fig:bt}.
Since a selector with memory \textit{remembers} which branch is running from one polling to the other,
the presence of new objects on the conveyor is ignored until the execution of the current skill succeeds.

\begin{figure}[t]
\centering
\includegraphics[width = 0.9\columnwidth]{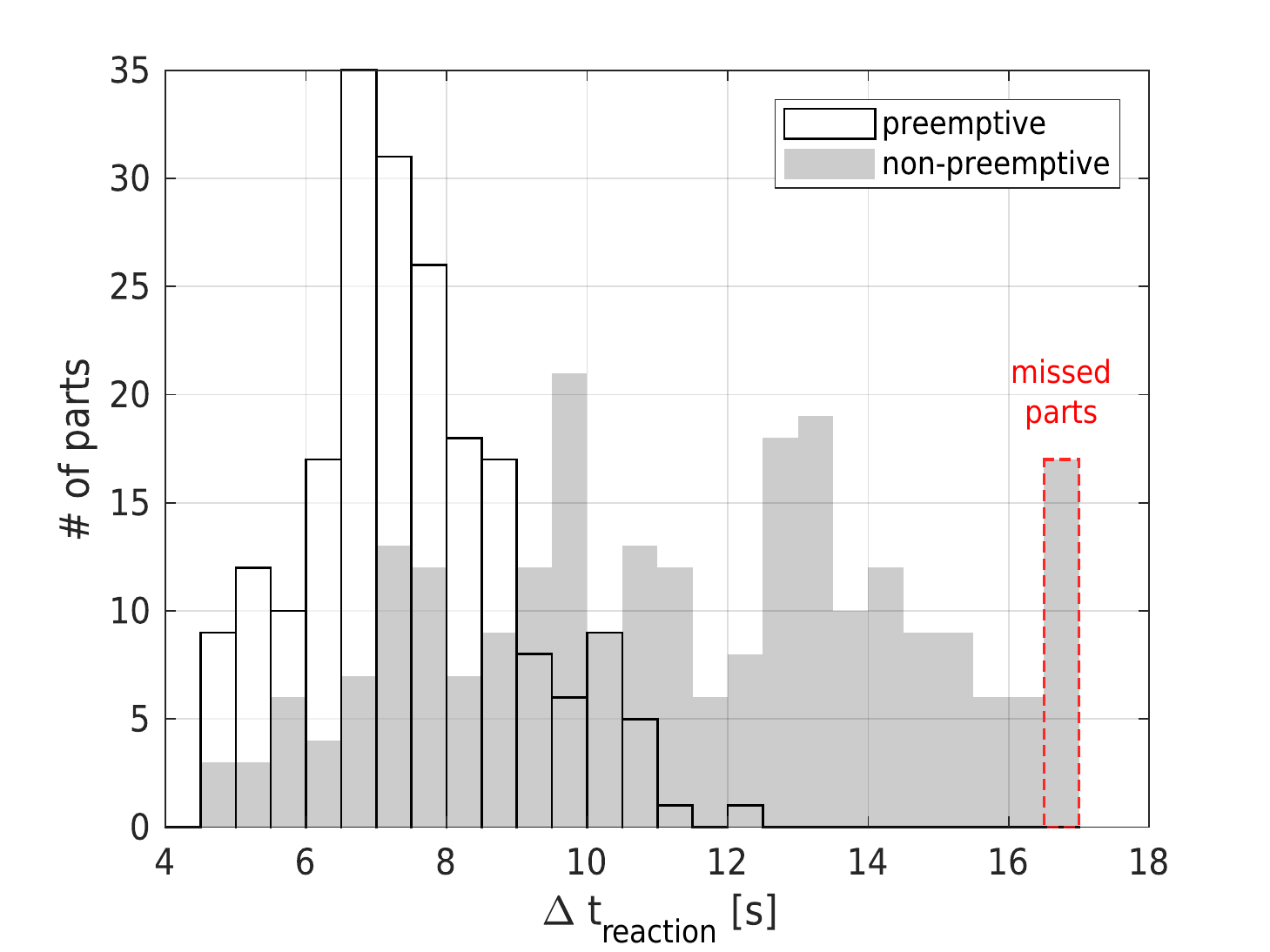}
\caption{Histogram of the distribution of $\Delta t_\text{reaction}$ for preemptive and non-preemptive skills.
$\Delta t_\text{reaction}$ corresponds to the time elapsed
between the detection of a part on the conveyor belt and the removal of this part.
The parts on the far right were missed by the robot and therefore exited the conveyor system.}
\label{fig:histogram}
\end{figure}

We evaluated our implementation against this baseline by conducting 80 trials in the environment of ARIAC\footnote{
The ROS package allowing to replicate these trials is available in open-source on GitHub:
\href{https://github.com/ScalABLE40/productive\_multitasking}{https://github.com/ScalABLE40/productive\_multitasking}}:
40 with the trick, and 40 without.
A trial consists of placing 18 parts on the delivery tray of the AGV
starting with 9 laying in the second bin (counting from the left-hand side)
and 9 others in the third bin.
More parts are being conveyed at regular time interval $\Delta T_a$ during the whole trial
and $\Delta T_a = 30 + 0.2 k$ with $k \in [40]$.
This experiment resulted in the execution of 3134 skills with 150 preemptions.
234 parts were picked up on the conveyor by non-preemptive skills,
and 205 by preemptive skills.
$\Delta t_\text{reaction}$ corresponds to the time elapsed between
the detection of a part on the conveyor belt and the removal of this part.
The time window for picking up detected parts on the conveyor belt is $\Delta t^* = 16 s$.
Fig.~\ref{fig:histogram} shows the distribution of $\Delta t_\text{reaction}$
for both preemptive and non-preemptive skills.
As it can be seen, preemptive skills allow to never miss a detected parts,
while non-preemptive ones result in 17 lost parts.
Also, the maximum value of $\Delta t_\text{reaction}$ for preemptive skills equals 75 \% of the allowed time window $\Delta t^*$.
Hence, the conveyor belt could be run approximately 33 \% faster,
or the length of the robot workspace covering the conveyor belt could be reduced of approximately 25 \%.

\subsection{Scalability of the BT-FSM model for Preemption}

The FSM of Fig.~\ref{fig:fsm},
\textit{with} the possibility of triggering a preemption on the placing sequence,
has also been used in the \textit{complete} scenario of ARIAC.
The ARIAC scenario also involves another robot, order updates, gripper failures,
collaboration between the robots and detection of faulty parts.
With this, we were able to verify that our approach can scale up.
The BT consists of 16 control flow nodes (4 on the top level plus 6 per robot) and 
30 relatively computationally intensive execution nodes (8 on the top level plus 11 per robot).
The FSM supports all the possible transitions between the skills,
and the BT can exploit all these transitions.
However, a discussion of the full set of ARIAC experiments is beyond the scope of this paper.

%% file: tex/discussion.tex
\section{Discussion and Conclusion}
\label{sec:discussion}




Within the ARIAC competition we have attempted to use classic BTs and FSM to model the tasks. 
However, due to the task complexity and the possible preemptions, both models proved limited: FSM easily grew complex and lost their readability, while BTs proved challenging to design a whole solution without embedding a fixed state memory for some transitions. 
The BT-FSM model has not only proven to be much simpler to use and more structured for the large application, but using BT and FSM in combination we were able to retain their respective advantages.
Combining BTs and FSMs gives the best out of the two paradigms: BTs model reactive transitions, while FSMs keep the state memory without any need to define conditions with a possible complex semantic.
Incorporating FSMs in BTs provides a more systematic alternative to other solutions found in the literature, as for example the star operator introduced by Marzianotto et Al. \cite{2014Marzinotto-BT}.
In our experiments, the BT-FSM model has proven to be more maintainable, and debugging was considerably easier. 
Like the BTs before, the BT-FSM model that we introduced here was also motivated by know-how from the video gaming industry.
A more systematic analysis and exploration of these features is, however, beyond the scope of this paper.

In terms of preemption we employed the BT-FSM framework in the environment of ARIAC
and showed the benefits of preemptive skills w.r.t. non-preemptive ones.
Our implementation resulted in a robot capable of reacting instantaneously
to the occurrences of non-deterministic events\textemdash
similarly to \textit{reflexes}, but without short-circuiting any deliberative loop.
Future work will include the application of complexity metrics to BTS and FSMs
in order to determine in which cases one model is preferable over the other.

\section*{Acknowledgment}

The authors would like to thank the National Institute of Standards and Technology (NIST)
and the Open Source Robotics Foundation (OSRF)
for organizing the Agile Robotics for Industrial Automation Competition (ARIAC).